\documentclass[10pt,twocolumn,letterpaper]{article}
\usepackage{textcomp}
\usepackage{amsmath}
\usepackage{amssymb}
\usepackage{times}
\usepackage{epsfig}
\usepackage{bm}
\usepackage{xargs}
\usepackage{multirow}
\usepackage{amsfonts}
\usepackage{color}
\usepackage{booktabs}

\graphicspath{{res/}}

\newcommandx\includeImageLineWidth[2][1=1.0]{\includegraphics[width=#1\linewidth]{#2}}
\newcommand{\ba}{{\bm{a}}}

\newcommand{\bx}{{\bm{x}}}

\newcommand{\bW}{{\bm{W}}}

\usepackage{array}
\newcommand{\PreserveBackslash}[1]{\let\temp=\\#1\let\\=\temp}
\newcolumntype{C}[1]{>{\PreserveBackslash\centering}p{#1}}
\newcolumntype{R}[1]{>{\PreserveBackslash\raggedleft}p{#1}}
\newcolumntype{L}[1]{>{\PreserveBackslash\raggedright}p{#1}}

\usepackage{iccv}
\usepackage{times}
\usepackage{epstopdf}
\usepackage{graphicx}
\usepackage{amsmath}
\usepackage{amssymb}
\usepackage{subfigure}
\usepackage{algorithm}
\usepackage{algorithmic}

\usepackage[breaklinks=true,bookmarks=false]{hyperref}

\iccvfinalcopy 

\def\iccvPaperID{****} 

\ificcvfinal\fi
\begin{document}

\title{AM-LFS: AutoML for Loss Function Search}

\author{Chuming Li\textsuperscript{1}\thanks{equal contribution}, Xin Yuan\textsuperscript{1}\footnotemark[1],
Chen Lin\textsuperscript{1}\footnotemark[1], Minghao Guo\textsuperscript{1}, Wei Wu\textsuperscript{1}, Junjie Yan\textsuperscript{1}, Wanli Ouyang\textsuperscript{2}\\
\textsuperscript{1}SenseTime Group Limited \\
\textsuperscript{2}The University of Sydney \\
{\tt\small \{lichuming,yuanxin,linchen,guominghao,wuwei,yanjunjie\}@sensetime.com; wanli.ouyang@sydney.edu.au}
}

\maketitle
\ificcvfinal\thispagestyle{empty}\fi

\begin{abstract}
Designing an effective loss function plays an important role in visual analysis. Most existing loss function designs rely on hand-crafted heuristics that require  domain  experts  to  explore  the  large  design  space, which is usually sub-optimal and time-consuming. In this paper, we propose AutoML for Loss Function Search (AM-LFS) which leverages REINFORCE to search loss functions during the training process. The  key  contribution of this work is the design of search space which can guarantee the generalization and transferability on different vision tasks by including a bunch of existing prevailing loss functions in a unified formulation. We also propose an efficient optimization framework which can dynamically optimize the parameters of loss function's distribution during training. Extensive experimental results on four benchmark datasets show that, without any tricks, our method outperforms existing hand-crafted loss functions in various computer vision tasks.
\end{abstract}
 
\section{Introduction}

\begin{figure}[t]
\centerline{\includegraphics[width=1.0\columnwidth]{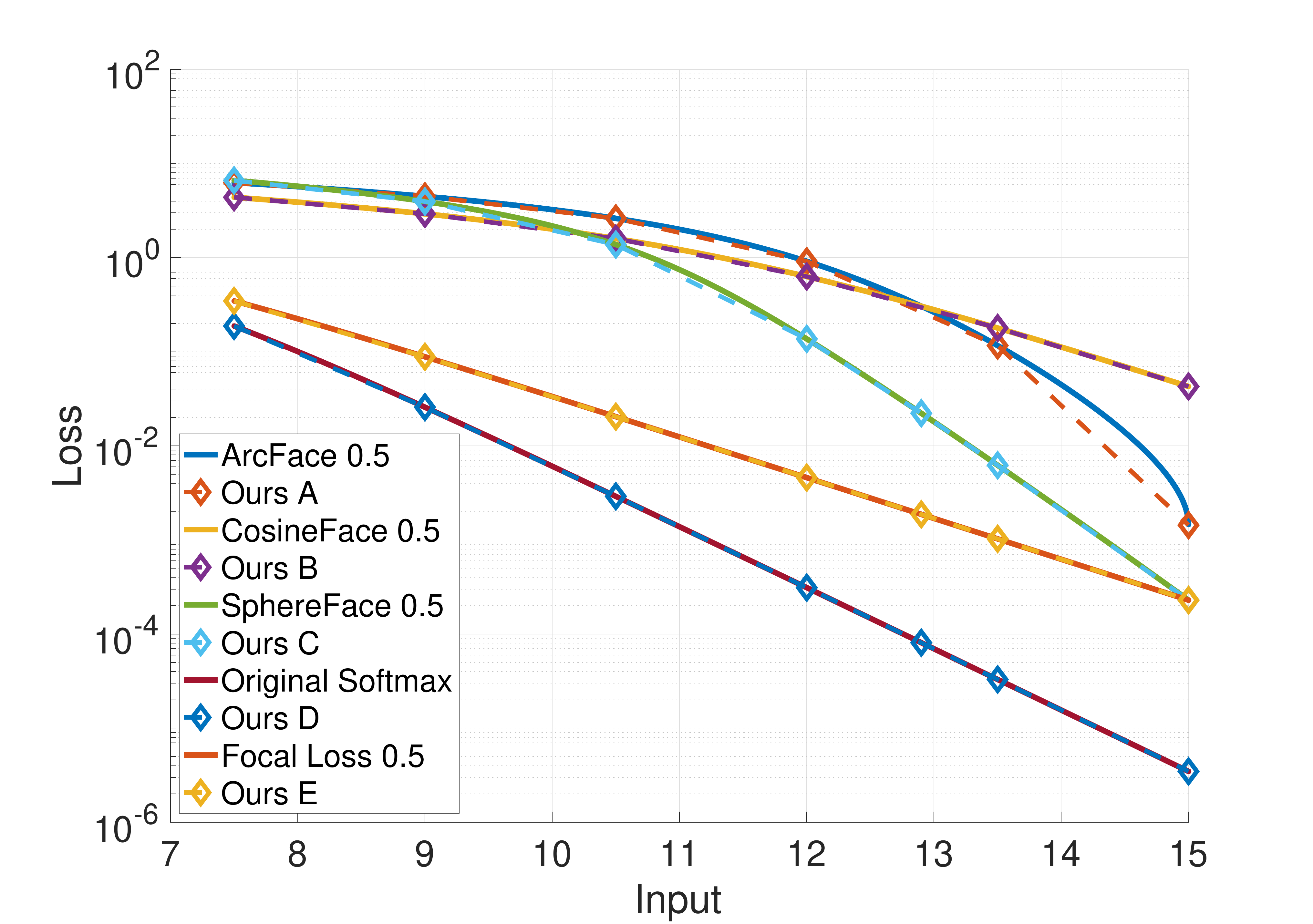}}
\caption{The motivation of the proposed loss function search space. 
The figure shows that the candidate loss functions in our search space (dotted lines) can well approximate the according existing loss functions (solid lines). The x-axis indicates the loss input and the y-axis indicates the output loss values in log-scale.
}
\label{fig:idea}
\end{figure}

Convolutional neural networks have significantly boosted the performance of a variety of visual analysis tasks, such as image classification~\cite{krizhevsky2012imagenet, DBLP:journals/corr/SimonyanZ14a, he2016deep}, face recognition~\cite{DBLP:conf/cvpr/LiuWYLRS17, DBLP:journals/spl/WangCLL18, DBLP:journals/corr/abs-1801-07698}, person re-identification~\cite{DBLP:journals/corr/abs-1811-11405, DBLP:conf/mm/WangYCLZ18, DBLP:conf/eccv/SuhWTML18, DBLP:journals/corr/abs-1807-00537} and object detection~\cite{DBLP:conf/iccv/Girshick15, DBLP:conf/nips/RenHGS15} in recent years due to its high capacity in learning discriminative features.
Aside from developing features from deeper networks to get better performance, better loss functions  have  also  been  proven  to  be  effective  on  improving the performance of the computer vision frameworks in most recent works~\cite{DBLP:conf/cvpr/LiuWYLRS17, DBLP:conf/iccv/LinGGHD17}.

Conventional CNN-based vision frameworks usually apply a widely-used softmax loss to high level features.
L-softmax~\cite{DBLP:conf/icml/LiuWYY16} is a variant of softmax loss which added multiplicative angular to each class to improve feature discrimination in classification and verification tasks.
~\cite{DBLP:conf/cvpr/LiuWYLRS17} introduced A-softmax by applying L-softmax~\cite{DBLP:conf/icml/LiuWYY16} to face recognition task with weights normalization.
~\cite{DBLP:conf/cvpr/WangWZJGZL018, DBLP:journals/spl/WangCLL18} moved the angular margin into cosine space to overcome the optimization
difficulty of~\cite{DBLP:conf/cvpr/LiuWYLRS17} and achieved state-of-the-art performance.
~\cite{DBLP:journals/corr/abs-1801-07698} can obtain more discriminative deep features for face recognition by incorporating the additive angular margin. In addition to the above margin-based softmax loss functions,
focal loss~\cite{DBLP:conf/iccv/LinGGHD17} is another variant of softmax loss which was proposed to adopt a re-weighting scheme to address the data imbalance problems in object detection.
While these methods improve performance over the traditional softmax loss, they still come with some limitations:
(1) Most existing methods rely on hand-crafted heuristics that require great efforts from domain experts  to  explore  the  large  design  space, which is usually sub-optimal and time-consuming. (2) These methods are usually task-specific, which may lack transferability when applied to other vision tasks.
By utilizing AutoML methods for exploration in a well-designed loss function search space, a generic solution can be proposed to further improve the performance.

In this paper, we propose AutoML for Loss Function Search (AM-LFS) method from a hyper-parameter optimization perspective.
Based on the analysis of existing modification on the loss functions,
we design a novel and effective search space, which is illustrated in Figure~\ref{fig:idea} and formulate hyper-parameters of loss function as a parameterized probability distribution for sampling.
The proposed search space include a bunch of popular loss designs, whose sampled candidate loss functions can adjust the gradients of examples at different difficulty levels and balance the significance of intra-class distance and inter-class distance during training. 
We further propose a bilevel framework which allows the parameterized distribution to be optimized simultaneously with the network parameters. 
In this bilevel setting, the inner objective is the minimization of sampled loss w.r.t network parameters, while the outer objective is the maximization of rewards (e.g. accuracy or mAP) w.r.t loss function distribution. 
After the AM-LFS training finishes, the network parameters can be directly deployed for evaluation and then get rid of the heavily re-training steps.
Furthermore, since our method is based on the loss layer without modification on  the network architecture of specific tasks, it can be easily applied to off-the-shelf modern classification and verification frameworks.
We summarize the contributions of this work as follows:

(1) We provide an analysis based on the existing loss function design and propose a novel and effective search space which can guarantee the generalization and transferability on different vision tasks.

(2) We propose an efficient optimization framework which can dynamically optimize the distribution for sampling of the loss functions.

(3) The proposed approach advances the performance of the state-of-the-art methods on popular classification, face and person re-id databases including CIFAR-10, MegaFace, Market-1501 and DukeMTMC-reID.

\section{Related Work}
\subsection{Loss Function}
Loss function plays an important role in deep feature learning of various computer vision tasks.
Softmax loss is a widely-used loss for CNN-based vision frameworks.
A large margin Softmax (L-Softmax)~\cite{DBLP:conf/icml/LiuWYY16} modified softmax loss by adding multiplicative angular constraints to each identity to improve feature discrimination in classification and verification tasks.
SphereFace~\cite{DBLP:conf/cvpr/LiuWYLRS17} applies L-Softmax to deep face recognition with weights normalization.
CosineFace~\cite{DBLP:conf/cvpr/WangWZJGZL018, DBLP:journals/spl/WangCLL18} and ArcFace~\cite{DBLP:journals/corr/abs-1801-07698} can achieve the state-of-the-art performance on the MegaFace with more discriminative deep features by incorporating the cosine margin and additive angular margin, respectively.
SphereReID~\cite{DBLP:journals/corr/abs-1807-00537} adopted the sphere softmax and trained the model end-to-end to achieve the state-of-the-art results on the challenging person reid datasets.
For object detection, focal loss~\cite{DBLP:conf/iccv/LinGGHD17} and gradient harmonized detector~\cite{DBLP:journals/corr/abs-1811-05181} adopt a re-weighting scheme to address the class imbalance problem. However, noisy labels can lead to misleading gradient, which may be amplified by gradient re-weighting schemes and cause a training failure.

\begin{figure*}[tb]
\centerline{\includegraphics[width=2.0\columnwidth]{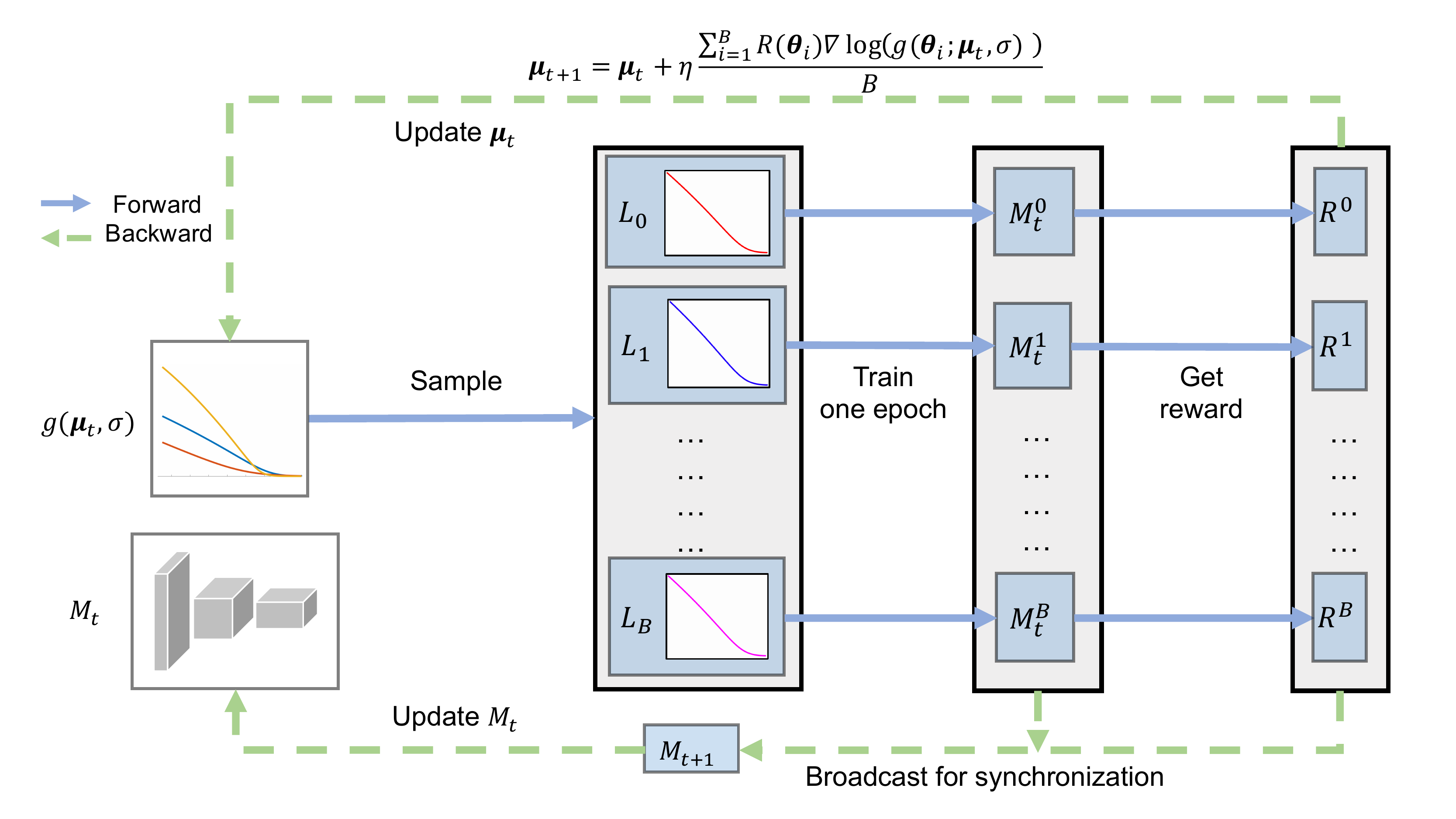}}
\caption{The bilevel optimization framework of our proposed AM-LFS approach. In this bilevel setting, the inner objective is the minimization of sampled loss w.r.t network parameters, while the outer objective is the maximization of rewards (e.g. accuracy or mAP) w.r.t loss function distribution. After each train epoch, we broadcast the model parameters with the highest reward to each sample for synchronization.}
\label{fig:framework}
\end{figure*}

\subsection{AutoML}
AutoML was proposed as an AI-based solution to the challenging tasks by offering faster solutions creation and models outperforming those designed by hand.
Recent works of automatically searching neural network architectures (NAS) has greatly improve the performance of neural networks.
NAS utilizes reinforcement learning~\cite{zoph2017learning,zoph2016neural} and genetic algorithms~\cite{real2018regularized,xie2017genetic,saxena2016convolutional} to search the transferable network blocks whose performance surpasses many manually designed architectures.
However, all the above methods require massive computation during the search, particularly thousands of GPU days. 
Recent efforts such as~\cite{liu2018darts, luo2018neural, pham2018efficient}, utilized several techniques trying to reduce the search cost.
\cite{liu2018darts} is a differential based method which utilized a bilevel optimization procedure for the jointly training of the real-valued architecture parameters and the model parameters.
Several methods attempt to automatically search architectures with fast inference speed by explicitly taking the inference latency as a constrain \cite{tan2018mnasnet, DBLP:journals/corr/abs-1812-00332} or implicitly encoding the topology information \cite{guo2018irlas}. 
In addition to the network architecture search, ~\cite{DBLP:conf/eccv/HeLLWLH18} utilized AutoML techniques for effective model compression, where the pruning rate was automatically decided by reinforcement learning.
\cite{DBLP:conf/nips/WuTXFQLL18} utilized a teacher model inspired by \cite{DBLP:journals/corr/abs-1805-03643} to guide the student model training with dynamic loss function. However, how to design a generic search space to various domains of tasks and an efficient optimization framework remain an open problem.

\section{Approach}
In this section, we first revisit several loss function designs from a novel perspective and then analyze their influence on the training procedure and reformulating them in a unified expression.
We hence propose a novel search space on the basis of the unified expression to include the good properties of existing popular loss function designs.
We also propose an optimization framework by utilizing AutoML methods for efficient loss function search during the whole training process, which is illustrated in Figure~\ref{fig:framework}.
\subsection{Revisiting Loss Function Design}
\textbf{Softmax Loss:} The most widely used softmax loss can be written as
\begin{eqnarray}
    L = \frac{1}{N}\sum_i{L_i} = \frac{1}{N}\sum_i{-log\left(\frac{e^{f_{y_i}}}{\sum_j{e^{f_j}}}\right)}, 
\end{eqnarray}
where $\bx_i$ and $y_i$ is the $i$-th input feature and the label, respectively. $f_j$ denotes the $j$-th element ($j \in [1,C]$, where $C$ is the number of classes) of the vector of scores $\boldsymbol{f}$, $N$ is the length of training set. $\boldsymbol {f}$ is usually the activation of a fully connected layer $\bW$. We further denote $f_{j}$ as $\bW_{j}^{T}\bx_i $, where $\bW_{j}$ is the $j$-th column of $\bW$.
Hence $f_j$ can be formulated as:
\begin{eqnarray}
f_j = \left\|\bW_j\right\|\left\|\bx_i\right\|cos\left( \theta_j \right),
\end{eqnarray}
where $\theta_j(0\leq\theta_j\leq\pi)$ is the angle between the vector $\bW_j$ and $\bx_i$, $\left\|\bW_j\right\|$ and $\left\|\bx_i\right\|$ are $L_2$ norm of $\bW_j$ and $\bx_i$. Note that $\left\|\bW_j\right\|=1$ or $\left\|\bx_i\right\|=1$ if $\bW_j$ or $\bx_i$ is normalized. Thus the original softmax loss can be rewritten as:
\begin{eqnarray}\label{eq:softmax2}
    L_i = -log\left(  \frac{e^{\left\|\bW_{y_i}\right\|\left\|\bx_i\right\|cos\left( \theta_{y_i} \right)}}{\sum_j{e^{\left\|\bW_j\right\|\left\|\bx_i\right\|cos\left( \theta_j \right)}}}  \right)
\end{eqnarray}
We can easily obtain several variants of the original softmax loss such as margin-based softmax loss and focal loss by inserting transforms into Eq.~\ref{eq:softmax2}.

\textbf{Margin-based Softmax Loss:} The family of margin-based loss functions can be obtained by inserting a continuously differentiable transform function $t\left(\right)$ between the the norm $\left\|\bW_{y_i}\right\|\left\|\bx_i\right\|$ and $cos\left( \theta_{y_i} \right)$, which can be written as:
\begin{eqnarray}
L_i^t =& -log\left(  \frac{e^{\left\|\bW_{y_i}\right\|\left\|\bx_i\right\|t\left(cos\left( \theta_{y_i} \right)\right)}}{e^{\left\|\bW_{y_i}\right\|\left\|\bx_i\right\|t\left(cos\left( \theta_{y_i} \right)\right)}+\sum_{j\not=y_i}{e^{\left\|\bW_j\right\|\left\|\bx_i\right\|cos\left( \theta_j \right)}}}  \right),
\end{eqnarray}
where $t$ is used to distinguish margin-based softmax loss functions with different transforms. 
We also list several transforms including L-softmax, A-softmax, ArcFace and their corresponding expressions in Table.~\ref{tab:express},

\begin{table}[tb]\label{tab:express}
\begin{center}
\caption{The expressions of some existing transforms including L-softmax, A-softmax and ArcFace.}
\label{tab:megaface}
\begin{tabular}{lcr}
\toprule
&\multicolumn{1}{c}{{transform}}
&\multicolumn{1}{c}{expression} \\
\midrule
&L-softmax~\cite{DBLP:conf/icml/LiuWYY16} & $ t(x) = cos(m\cdot arccos(x)) $ \\
&A-softmax~\cite{DBLP:conf/cvpr/LiuWYLRS17} & $ t(x) = x + m $  \\
&ArcFace~\cite{DBLP:journals/corr/abs-1801-07698} & $ t(x) = cos( arccos(x)+m ) $ \\
 \bottomrule
\end{tabular}
\end{center}
\end{table}

\textbf{Focal Loss:} This is also a variant which can be derived from softmax loss by adding transform function at another location, which can be described as
\begin{equation}
L_i^t = -\tau(log(p_{y_i}))
\end{equation}
\begin{equation}
\tau(x) = x(1-e^x)^\alpha.
\end{equation}

\subsection{Analysis of Loss Function}
In this section, we first discuss the impacts of margin-based softmax loss functions on the training process from a new perspective based on the analysis of the relative significance of intra-class distance and inter-class distance. 
We define the inter-class distance $d_j$ as the distance between the feature $\bx_i$ and the class center $\bW_{j}$. The intra-class distance $d_{y_i}$ can be defined in a similar way. For simplification, we assume that $\bW_{j}$ and $\bx_i$ are normalized, which means $\left\|\bW_{j}\right\| = \left\|\bx_i\right\| = 1$ and $f_j = cos(\theta_j)$. Under this assumption, the relationship between $f_j$ and $d_j$ can be described as follows:
\begin{eqnarray}
d_j^2 = (\bx_i - \bW_j)^2 = 2 - 2f_j. 
\end{eqnarray}
As a result, we can analyze the impact of margin-based softmax loss on $f_{y_i}$ and $f_j$ instead. The impact can be calculated as the norm of gradient passed to the activation layer $\boldsymbol{f}$. Specifically, the gradients of the loss layer with respect to $f_{y_i}$ and $f_j$ are:
\begin{eqnarray}
&||\frac{\partial L_i^t}{\partial f_{y_i}}|| = (1-p_{y_i}^t)t'(f_{y_i}), \\
&||\frac{\partial L_i^t}{\partial f_j}|| = p_j^t,
\end{eqnarray}
where
\begin{eqnarray}
p_{y_i}^t = \frac{e^{\left\|\bW_{y_i}\right\|\left\|\bx_i\right\|t\left(cos\left( \theta_{y_i} \right)\right)}}{e^{\left\|\bW_{y_i}\right\|\left\|\bx_i\right\|t\left(cos\left( \theta_{y_i} \right)\right)}+\sum_{j\not=y_i}{e^{\left\|\bW_j\right\|\left\|\bx_i\right\|cos\left( \theta_j \right)}}}, \\
p_j^t = \frac{e^{\left\|\bW_j\right\|\left\|\bx_i\right\|cos\left( \theta_j \right)}}{e^{\left\|\bW_{y_i}\right\|\left\|\bx_i\right\|t\left(cos\left( \theta_{y_i} \right)\right)}+\sum_{j\not=y_i}{e^{\left\|W_j\right\|\left\|\bx_i\right\|cos\left( \theta_j \right)}}}
\end{eqnarray}
We further define \textbf{relative significance} of intra-class distance to inter-class distance as the \textbf{ratio} of norms of the gradients of $f_{y_i}$ and $f_j$ with respect to the margin-based softmax loss, which is described as follows:
\begin{eqnarray}
r_i^t= \frac{||\frac{\partial L_i^t}{\partial f_{y_i}}||}{||\frac{\partial L_i^t}{\partial f_j}||} = \frac{(1-p_{y_i}^t)}{p_j^t}t'(f_{y_i}), 
\end{eqnarray}
while with respect to the original softmax loss, this significance ratio is
\begin{eqnarray}
r_i^o= \frac{||\frac{\partial L_i^o}{\partial f_{y_i}}||}{||\frac{\partial L_i^o}{\partial f_j}||} = \frac{(1-p_{y_i}^o)}{p_j^o}.
\end{eqnarray}
Where $o$ is the identity transform.
The impact of margin-based loss on the relative significance of intra-class distance to inter-class distance can be calculated as the ratio of $r_i^t$ and $r_i$:
\begin{eqnarray}
\frac{r_i^t}{r_i^o} = \frac{\frac{(1-p_{y_i}^t)}{p_j^t}}{\frac{(1-p_{y_i}^o)}{p_j^o}}t'(f_{y_i}) &=& \frac{\frac{\sum_{t\not=y_i}{e^{\left\|W_t\right\|\left\|x_i\right\|cos\left( \theta_t \right)}}}{e^{\left\|W_j\right\|\left\|x_i\right\|cos\left( \theta_j \right)}}}{\frac{\sum_{t\not=y_i}{e^{\left\|W_t\right\|\left\|x_i\right\|cos\left( \theta_t \right)}}}{e^{\left\|W_j\right\|\left\|x_i\right\|cos\left( \theta_j \right)}}}t'(f_{y_i}) \nonumber \\
&=& t'(f_{y_i})
\end{eqnarray}
Here we conclude that margin-based softmax loss layer mainly functions as a controller to change the relative significance of intra-class distance to inter-class distance by its derivative $t'(f_{y_i})$ during the training process.
In addition to the margin-based softmax loss, we also analyze the impact of the focal loss.
The gradient of focal loss with respect to the activation $\boldsymbol f$ equals to that of original softmax loss multiply $\tau'(log(p_{y_i})$.
This gradient leads to a totally different yet very effective impacts on the training process, which monotonically decreases with the log-likehood and help balance samples at different levels of difficulty.

\subsection{Search Space}
Based on the analysis above, we can insert two transforms $\tau$ and $t$ into the original softmax loss to generate loss functions with a unified formulation, which have both capabilities of balancing (1) intra-class distance and inter-class distance, (2) samples at different levels of difficulty. 
The unified formulation can be written as:
\begin{eqnarray}\label{eq:space1}
L_i^{\tau,t} = -\tau(log\left( p_{y_i}^t \right)),
\end{eqnarray}
where $\tau$ and $t$ are any function with positive gradient.
To ensure $\tau$ has a bounded definition domain $[0,1]$ hence reduce the complexity of searching in this space, we exchange $\tau$ and $log$,
\begin{eqnarray}\label{eq:space2}
L_i^{\tau,t} = -log(\tau\left( p_{y_i}^t \right)).
\end{eqnarray}
We prove that these two search space defined in Eq.~\ref{eq:space1} and Eq.~\ref{eq:space2} are equivalent:
for any $\tau_1(x)$ in Eq.~\ref{eq:space1}, we can get the same loss function by simply setting $\tau_2(x) = e^{\tau_1(log(x))}$ in Eq.~\ref{eq:space2}.

Our search space is formulated by the choices of $\tau$ and $t$, whose impacts on the training procedure are decided by derivatives $\tau'$ and $t'$ according to the analysis above.
As a result, we simply set the candidate set as piecewise linear functions that evenly divide the definition domain, which ensures independent slopes and bias within each interval. Take the function $t$ as an example:
\begin{eqnarray}
t(x) = a_i^t x+b_i^t, x \in [\zeta_i^t , \zeta_{i+1}^t],
\end{eqnarray}
where $\boldsymbol{\zeta}^t=[\zeta_0, ... \zeta_M]$($M$ is the number of intervals) are the end points of the intervals, and $\zeta_{i+1}^t - \zeta_i^t = (\zeta_{M}^t - \zeta_0^t)/M$ for $i \in [0, M-1]$. 

We also analyze the effectiveness of the components in our search space.
Samples in different intervals are at different levels of difficulty. For example, larger-value intervals contain easier samples with smaller intra-class distance $d_{y_i}$. 
Since $t'$ denotes the relative significance between intra-class and inter-class distance, we assign each interval with independent slope $t' = a_i^t$ to ensure the loss function can independently balance the significance of intra-class and inter-class distance at different levels of difficulty.
Similarly, $\tau$'s candidate set is the described by $a_i^{\tau}$, $b_i^{\tau}$ and ${\zeta_i^{\tau}}$, which balances the significance of samples at different difficulty levels.
The biases $b_i^t$ and $b_i^{\tau}$ guarantees the independence of each interval from previous intervals. 
We set $\bm{\zeta}^t$ and $\bm{\zeta}^{\tau}$ as constant values which evenly divide definition domains into $M$ intervals. We define $\bm{\theta} = [{\bm{a}^t}^T,{\bm{b}^t}^T,{\bm{a}^{\tau}}^T,{\bm{b}^{\tau}}^T]^T$. Given $\bm{\zeta}^t$, $\bm{\zeta}^{\tau}$, the loss function $L^{\bm{\theta}}$ = $L^{t,\tau}$ can decided merely by $\bm{\theta}$.
Thus, our search space can be parameterized by $\mathcal{L}=\{L^{\bm{\theta}}\}$.

\subsection{Optimization}

Suppose we have a network model $M_{\bm{\omega}}$ parameterized by $\bm{\omega}$, train set $\mathcal{D}_{t}=\{(\bx_i,y_i)\}_{i=1}^{n}$ and validation set $\mathcal{D}_v$, our target of loss function search is to maximize the model $M_{\bm{\omega}}$'s rewards $r(M_{\bm{\omega}}; \mathcal{D}_v)$ (e.g. accuracy or mAP) on validation set $D_v$ with respect to $\bm{\theta} = [{\ba^t}^T,{\bm{b}^t}^T,{\bm{a}^{\tau}}^T,{\bm{b}^{\tau}}^T]^T$, and the model $M_{\bm{\omega}}$ is obtained by minimizing the search loss:
\begin{eqnarray}\label{object}
\begin{aligned}
& \mathop{\max}\limits_{\bm{\theta}} R(\bm{\theta}) = r{(M_{\bm{\omega}^*(\bm{\theta})}, \mathcal{D}_v)} \\
{\rm{s.t.}} \quad &{\bm{\omega}^*(\bm{\theta})}  = \mathop{\arg}\mathop{\min}\limits_{\bm{\omega}}\sum_{(x,y)\in {\mathcal{D}_{t}}} {L^{\bm{\theta}}}(M_{\bm{\omega}}(x),y),
\end{aligned}
\end{eqnarray}
This refers to a standard bilevel optimization problem~\cite{DBLP:journals/anor/ColsonMS07} where loss function parameters $\bm{\theta}$ are regarded as hyper-parameters. We trained model parameters $\bm{\omega}$ which minimize the training loss $L^{\bm{\theta}}$ at the inner level, while seeking a good loss function hyper-parameters $\bm{\theta}$ which results in a model parameter $\bm{\omega}^*$ that maximizes the reward on the validation set $\mathcal{D}_v$ at the outer level.

To solve this problem, we propose an hyper-parameter optimization method which samples $B$ hyper-parameters $\{\bm{\theta_1},...\bm{\theta_B}\}$ from a distribution at each training epoch and use them to train the current model.
In our AM-LFS, we model these hyper-parameters $\bm{\theta}$ as independent Gaussian distributions, described by 
\begin{equation}
\bm{\theta} \sim \mathcal{N}(\boldsymbol{\mu},\sigma I).
\end{equation}
After training for one epoch, $B$ models are generated and the rewards of these models are used to update the distribution of hyper-parameters by REINFORCE~\cite{DBLP:journals/ml/Williams92} as follows, 
\begin{eqnarray}\label{reinfoce_mu}
\boldsymbol{\mu_{t+1}} = \boldsymbol{\mu_{t}} + \eta \frac{1}{B}\sum_{i=1}^{B}R(\bm{\theta_i})\nabla_{\bm{\theta}} log(g(\bm{\theta_i};\boldsymbol{\mu_t},\bm{\sigma}))
\end{eqnarray}
, where $g(\bm{\theta};\boldsymbol{\mu},{\sigma})$ is PDF of Gaussian distribution.
The model with the highest score is used in next epoch. At last, when the training converges, we direct take the model with the highest score $r{(M_{\bm{w}^*(\bm{\theta})}, \mathcal{D}_v)}$ as the final model without any retraining.
To simplify the problem, we fix ${\sigma}$ as constant and optimize over $\boldsymbol{\mu}$. The training procedure of our AM-LFS is summarized in Algorithm~\ref{alg:AM-LFS}.

\begin{algorithm}[tb]
   \caption{:AM-LFS}
   \label{alg:AM-LFS}
\begin{algorithmic}
   \STATE {\bfseries Input:} Initialized model $M_{\bm{\omega}_0}$, initialized distribution $\bm{\mu}_0$, total training epochs $T$, distribution learning rate $\eta$
      \STATE {\bfseries Output:} Final model $M_{\bm{\omega}_T}$
    \FOR{$t=1$ {\bfseries to} $T$}
    \STATE Sample $B$ hyper-parameters $\bm{\theta}_1,...\bm{\theta}_B$ via distribution $\mathcal{N}(\boldsymbol{\mu}_t,\sigma I)$;
    \STATE  Train the model $M_{\bm{\omega}_t}$ for one epoch separately with the sampled hyper-parameters and get $M_{\bm{\omega}_t^1},...M_{\bm{\omega}_t^B}$;
    \STATE  Calculate the score $R(\bm{\theta}_1),...R(\bm{\theta}_B)$ 
    \STATE  Decide the index of model with highest score $i = \mathop{\arg}\mathop{\max}\limits_{j} R(\bm{\theta}_j)$;
    \STATE  Update $\bm{\mu}_{t+1}$ using Eq.~(\ref{reinfoce_mu})
    \STATE  Update $M_{\bm{\omega}_{t+1}} = M_{\bm{\omega}_t^i}$
    \ENDFOR
    \STATE return $M_{\bm{\omega}_T}$
\end{algorithmic}
\end{algorithm}

\section{Experiments}
We conducted experiments on four benchmarking datasets including CIFAR-10~\cite{krizhevsky2009learning}, MegaFace~\cite{DBLP:conf/cvpr/Kemelmacher-Shlizerman16}, Market-1501~\cite{DBLP:conf/iccv/ZhengSTWWT15} and DukeMTMC-reID~\cite{DBLP:conf/eccv/RistaniSZCT16,DBLP:conf/iccv/ZhengZY17} to show the effectiveness of our method on classification, face recognition and person re-identification tasks.
\subsection{Implementation Details}
Since our AM-LFS utilizes a bilevel optimization framework, our implementation settings can be divided into inner level and outer level.
In the inner level, 
for fair comparison, we kept all experimental settings such as warmup stage, learning rate, mini-batch size and learning rate decay consistent with the corresponding original baseline methods on the specific tasks.
Note that for the baseline methods with multi-branch loss, we only replace the softmax loss branch with AL-LFS while ignoring others. For example, in MGN~\cite{DBLP:conf/mm/WangYCLZ18}, we only apply AL-LFS to 8 softmax loss branches while keeping triplet loss unchanged. 
In the outer level, we optimized the distribution of loss functions in the search space, the gradients of distribution parameters were computed by REINFORCE algorithm with rewards from a fixed number of samples.  
The rewards are top-1 accuracy, rank 1 value and mAP for classification, face recognition and person re-identification, perspectively.
We normalized the rewards returned by each sample to zero mean and unit variance, which is set as the reward of each sample.
For all the datasets, the numbers of samples $B$ and intervals $M$ are set to 32 and 6, respectively. Note that we also conducted the investigation on these numbers in Section~\ref{sec:ablation}. 
For the distribution parameters, we used Adam optimizer with a learning rate of 0.05 and set $\sigma=0.2$.
Having finished the update of the distribution parameters, we broadcast the model parameters with the highest mAP to each sample for synchronization.
For all the datasets, each sampled model is trained with 2 Nvidia 1080TI GPUs, so a total of 64 GPUs are required.
\subsection{Datasets}
\textbf{Classification:}
The CIFAR-10 dataset consist of natural images with resolution $32 \times 32$.
CIFAR-10 consists of 60,000 images in 10 classes, with 6,000 images per class.
The train and test sets contain 50,000 and 10,000 images respectively.
On CIFAR datasets, we adopted a standard data augmentation scheme (shifting/mirroring) following~\cite{DBLP:journals/corr/LinCY13, DBLP:conf/eccv/HuangSLSW16}, and normalized the input data with channel means and standard deviations.

\textbf{Face Recognition:}
For face recognition, we set the CASIA-Webface~\cite{DBLP:journals/corr/YiLLL14a} as the training dataset and MegaFace as the testing dataset.
CASIA-Webface contains 494,414 training images from 10,575 identities.
MegaFace datasets are released as the largest public available testing benchmark, which aims at evaluating the performance of face recognition algorithms at the million scale of distractors.
To perform open-set evaluations, we carefully remove the overlapped identities between training dataset and the testing dataset.

\textbf{Person ReID:}
For person re-identification, we used Market-1501 and DukeMTMC-reID to evaluate our proposed AM-LFS method.
Market-1501 includes images of 1,501 persons captured from 6 different cameras. The pedestrians are cropped with bounding-boxes predicted by DPM detector. 
The whole dataset is divided into training set with 12,936 images of 751 persons and testing set with 3,368 query images and 19,732 gallery images of 750 persons.  In our experiments, we choose the single-query mode, where features are extracted from only 1 query image.
DukeMTMC-reID is a subset of Duke-MTMC for person re-identification which contains 36,411 annotated bounding box images of 1,812 different identities captured by eight high-resolution cameras. 
A total of 1,404 identities are observed by at least two cameras, and the remaining 408 identities are distractors. 
The training set contains 16,522 images of 702 identities and the testing set contains the other 702 identities.

\subsection{Results on CIFAR-10}
We demonstrate our method with ResNet-20~\cite{DBLP:conf/cvpr/HeZRS16} on the CIFAR-10 dataset~\cite{krizhevsky2009learning}.
We trained the model using the standard cross-entropy to obtain the original top-1 test error $8.75\%$.
Table~\ref{tab:cifar:noise} shows the classification results compared with standard cross-entropy (CE)~\cite{DBLP:journals/anor/BoerKMR05} and the dynamic loss by learning to teach (L2T-DLF)~\cite{DBLP:conf/nips/WuTXFQLL18} methods. As can be observed, among all three loss function methods, our AM-LFS helps ResNet-20 achieve the best performance of $6.92\%$ top-1 error rate.
We also see that the recently proposed L2T-DLF reduces the error rate of softmax loss by $1.12\%$ because L2T-DLF introduced the dynamic loss functions outputted via teacher model which can help to cultivate better student model.
Note that AM-LFS can further reduce the top-1 error rate of L2T-DLF by $0.71\%$, which attributes to more effective loss function search space and optimization strategy design in our method.

In addition to the conventional experiments, we also conducted the CIFAR-10 noisy label experiments, where labeled classes can randomly flip to any other label by a given noise ratio, to demonstrate AM-LFS has the property of data re-weighting during training. As shown in Table~\ref{tab:cifar:noise}, AM-LFS consistently outperforms the baseline softmax loss by $2.00\%$ and $2.32\%$ under the noise ratio of $10\%$ and $20\%$ respectively.

\begin{table}[tb]
\caption{Results on the dataset CIFAR-10 using ResNet-20, showing the ratio of noise label, the top-1 test error rate (\%) with standard cross entropy, L2T-DLF and AM-LFS.
}
\label{tab:cifar:noise}
\begin{tabular}{lcccr}
\toprule
&\multicolumn{1}{l}{{Noise ratio(\%)}}
&\multicolumn{1}{c}{CE~\cite{DBLP:journals/anor/BoerKMR05}} 
&\multicolumn{1}{c}{L2T-DLF~\cite{DBLP:conf/nips/WuTXFQLL18}} 
&\multicolumn{1}{c}{AM-LFS} \\
\midrule
 &0 &8.75 &7.63 &\textbf{6.92} \\
&10 &12.05 &--- &\textbf{10.05} \\
&20 &15.05 &--- &\textbf{12.73} \\
 \bottomrule
\end{tabular}
\end{table}
\subsection{Results on MegaFace}
We compare the proposed AM-LFS with three state-of-the-art loss function methods,
including SphereFace, CosineFace and ArcFace. 
We used a modified ResNet with 20 layers that is adapted to face recognition and followed the same experimental settings with~\cite{DBLP:journals/spl/WangCLL18} when training the model parameters at the inner level.
Table~\ref{tab:megaface} shows the MegaFace rank1@1e6 performance with various loss functions.
For SphereFace and CosineFace, we directly reported the results from the original paper. For ArcFace, we report their results by running the source codes provided by the authors to train the models by ourselves.
We can observe that our proposed AM-LFS outperforms all compared methods by $6.1\%$, $1.0\%$ and $1.1\%$, respectively.
The main reason is that the candidates sampled from our proposed search space can well approximate all these compared loss functions, which means their good properties can be sufficiently explored and utilized during the training phase.
Meanwhile, our optimization strategy enables that the dynamic loss can guide the model training of different epochs, which helps further boost the discrimination power.

\begin{table}[tb]
\begin{center}
\caption{Comparison with state-of-the-art loss functions on the MegaFace dataset using ResNet-20. For SphereFace and CosineFace, we directly reported the results from the original paper. For ArcFace, we report their results by running the source codes provided by their respective authors to train the models by ourselves following the same setting with CosineFace.}
\label{tab:megaface}
\begin{tabular}{lccr}
\toprule
&\multicolumn{1}{c}{{Method}}
&\multicolumn{1}{c}{MegaFace rank1@1e6 } \\
\midrule
&SphereFace~\cite{DBLP:conf/cvpr/LiuWYLRS17} &67.4 \\
&CosineFace~\cite{DBLP:journals/spl/WangCLL18} &72.5 \\
&ArcFace~\cite{DBLP:journals/corr/abs-1801-07698} &72.4 \\
&AM-LFS &\textbf{73.5} \\
 \bottomrule
\end{tabular}
\end{center}
\end{table}

\begin{table}[tb]
\begin{center}
\caption{Comparison with state-of-the-art methods on the Market-1501 dataset using ResNet 50 showing mAP, rank 1 and rank 5. RK refers to implementing re-ranking operation.
}
\label{tab:market1501}
\begin{tabular}{lcccr}
\toprule
&\multicolumn{1}{c}{{Methods}}
&\multicolumn{1}{c}{mAP} 
&\multicolumn{1}{c}{rank1} 
&\multicolumn{1}{c}{rank5} \\
\midrule
&MLFN~\cite{DBLP:conf/cvpr/ChangHX18} &74.3 &90.0 &--- \\
&HA-CNN~\cite{DBLP:conf/cvpr/LiZG18} &75.7 &91.2 &--- \\
&DuATM~\cite{DBLP:conf/cvpr/SiZLKKK018} &76.6 &91.4 &97.1 \\
&Part-aligned~\cite{DBLP:conf/eccv/SuhWTML18} &79.6 &91.7 &96.9 \\
&PCB~\cite{DBLP:conf/eccv/SunZYTW18} &77.4 &92.3 &97.2 \\
&SphereReID~\cite{DBLP:journals/corr/abs-1807-00537} &83.6 &94.4 &98.0\\
&SFT~\cite{DBLP:journals/corr/abs-1811-11405} &82.7 &93.4 &--- \\
&MGN~\cite{DBLP:conf/mm/WangYCLZ18} &86.9 &95.7 &--- \\
&MGN(RK)~\cite{DBLP:conf/mm/WangYCLZ18} &94.2 &\textbf{96.6} &--- \\
\midrule
&SphereReID+ours &84.4 &95.0 &98.1 \\
&SFT+ours &83.2 &93.6 &97.9 \\
&MGN+ours &88.1 &95.8 &\textbf{98.4} \\
&MGN(RK)+ours &\textbf{94.6} &96.1 &\textbf{98.4} \\
\bottomrule
\end{tabular}
\end{center}
\end{table}

\subsection{Results on Market-1501 and DukeMTMC-reID}
We demonstrate the effectiveness of our AM-LFS by applying it to some existing competitors including SphereReID, SFT and MGN.
We compare with current state-of-the-art methods on both datasets to show our performance advantage over the existing baseline.
We report the mean average precision (mAP) and the cumulative matching characteristics (CMC) at rank-1 and rank-5 on all the candidate datasets. 
On Market-1501 dataset, we only conduct experiments both in single-query mode.
The results on Market-1501 dataset and DukeMTMC-reID dataset are shown in Table~\ref{tab:market1501} and Table~\ref{tab:duke}, respectively.
We divide the results into two groups according to whether our AL-LFS is applied or not.
For Market-1501, MGN(RK)+AM-LFS outperforms best competitor MGN(RK) by $0.4\%$ in mAP.
We observe that MGN rank 1 exhibits a degradation ($0.5\%$) after adopting AM-LFS. The mainly reason is that AM-LFS utilized mAP-related reward for guidance, which may not always be consistent with the rank 1 value.
For DukeMTMC-reID, MGN(RK)+AM-LFS surpass all compared methods in terms of mAP and rank 1. 
We conclude that although the baseline models SphereReID, SFT and MGN already achieved very high results on both Market-1501 and DukeMTMC-ReID, applying AM-LFS to them can still help cultivate better model hence boost the performance consistently. 

\begin{table}[tb]
\begin{center}
\caption{Comparison with state-of-the-art methods on the DukeMTMC-ReID dataset using ResNet 50 showing mAP, rank 1 and rank 5. RK refers to implementing re-ranking operation. 
}
\label{tab:duke}
\begin{tabular}{lcccr}
\toprule
&\multicolumn{1}{c}{{Methods}}
&\multicolumn{1}{c}{mAP} 
&\multicolumn{1}{c}{rank1} 
&\multicolumn{1}{c}{rank5} \\
\midrule
&PSE~\cite{DBLP:conf/cvpr/SarfrazSES18} &62.0 &79.8 &89.7 \\
&MLFN~\cite{DBLP:conf/cvpr/ChangHX18} &62.8 &81.0 &--- \\
&HA-CNN~\cite{DBLP:conf/cvpr/LiZG18} &63.8 &80.5 &--- \\
&DuATM~\cite{DBLP:conf/cvpr/SiZLKKK018} &64.6 &81.8 &90.2 \\
&Part-aligned~\cite{DBLP:conf/eccv/SuhWTML18}  &69.3 &84.4 &92.2\\
&PCB+RPP~\cite{DBLP:conf/eccv/SunZYTW18}  &69.2 &83.3 &--- \\
&ShpereReID~\cite{DBLP:journals/corr/abs-1807-00537} &68.5 &83.9 &90.6\\
&SFT~\cite{DBLP:journals/corr/abs-1811-11405} &73.2 &86.9 &93.9 \\
&MGN~\cite{DBLP:conf/mm/WangYCLZ18} &78.4 &88.7 &---\\
&MGN(RK)~\cite{DBLP:conf/mm/WangYCLZ18} &88.6 &90.9 &--- \\
\midrule
&ShpereReID+ours &69.8 &84.3 &92.0\\
&SFT+ours &73.8 &87.0 &95.1\\
&MGN+ours &80.0 &89.9 &95.2\\
&MGN(RK)+ours &\textbf{90.1} &\textbf{92.4} &\textbf{95.7} \\
\bottomrule
\end{tabular}
\end{center}
\end{table}

\subsection{Ablation Study}\label{sec:ablation}

\textbf{Effectiveness of the components:}
We showed the importance of both components \textbf{search space} and \textbf{search strategy} by demonstrating (1) the proposed design of the loss function search space itself can lead to AM-LFS's strong empirical performance and (2) the AM-LFS's capability of dynamically learning the distributions of better loss functions hence boost the performance during the training process.
We trained the SphereReID model on Market-1501 by sampling candidates from the initial distribution while not updating this distribution.
At the convergence, the model has the mAP of $84.0\%$, which outperforms the original baseline by $0.4\%$.
We conducted this study for the proposed search space, which can guarantee a performance gain over the baseline model even under a guided random search setting. 
We further enabled the optimization of the distribution and obtained an additional performance gain of $0.4\%$.
We therefore conclude that both the design of the loss function search space and the appropriate optimization procedures are crucial for good performance.

\begin{table}[tb]
\begin{center}
\caption{Effects of the number of samples by setting $B$ as 4, 8, 16, 32 in terms of mAP on the Market-1501 dataset and DukeMTMC-reID dataset using our AM-LFS based on the SphereReID baseline model.}
\label{tab:samples}
\begin{tabular}{lccccr}
\toprule
&\multicolumn{1}{c}{{Method}}
&\multicolumn{1}{c}{B=4} 
&\multicolumn{1}{c}{B=8} 
&\multicolumn{1}{c}{B=16} 
&\multicolumn{1}{c}{B=32}  \\
\midrule
&Market-1501 &83.6 &83.8 &84.2 &84.4 \\
&DukeMTMC-reID &68.4 &68.9 &69.7 &69.8 \\
 \bottomrule
\end{tabular}
\end{center}
\end{table}

\textbf{Investigation on samples:}
We study the effects of the number of samples in the optimization procedure by changing the parameter $B$ in AM-LFS.
Note that it costs more computation resources (GPUs) to train a minibatch of data as $B$ increases.
We report the performance results of different $B$ values selected from $\{4, 8, 16, 32\}$ in Table.~\ref{tab:samples} in terms of mAP on the Market-1501 and DukeMTMC-reID based on SphereReID.
The results show that when $B$ is small, the performance degrades because efficient gradients cannot be obtained without enough samples.
We also observe that the performance exhibits saturation when we keep enlarging $B$.
For a tradeoff of the performance and the training efficiency, we choose to fix $B$ as 32 during training.

\begin{table}[tb]
\begin{center}
\caption{Effects of the number of intervals by setting $M$ as 3, 6, 10 in terms of mAP on the Market-1501 dataset and DukeMTMC-reID dataset using our AM-LFS based on the SphereReID baseline model.}
\label{tab:interval}
\begin{tabular}{lccccr}
\toprule
&\multicolumn{1}{c}{{Dataset}}
&\multicolumn{1}{c}{M=3} 
&\multicolumn{1}{c}{M=6} 
&\multicolumn{1}{c}{M=10} \\
\midrule
&Market-1501 &83.8 &84.4 &84.2 \\
&DukeMTMC-reID &68.6 &69.8 &69.5 \\
 \bottomrule
\end{tabular}
\end{center}
\end{table}

\textbf{Investigation on intervals:}
We study the effects of the the number of intervals in the search space by changing the interval parameter $M$ in AM-LFS.
According to our design of the search space, samples at different levels of  difficulty are assigned to specific intervals, which enables a dynamic tradeoff between the intra-class distance and inter-class distance.
Table.~\ref{tab:interval} shows mAP performance results with respect to $M$ on the Market-1501 dataset and DukeMTMC-reID dataset.
When we set the interval number as a small number ($M = 3$), the mAP exhibits a low value because the intervals are not enough to handle all levels of hard examples during training.
The network is hard to train and suffers from an accuracy degradation when we set a large interval number ($M = 10$) because the excessive distribution parameters are hard to optimize.
We also observe that the best performance is achieved at a moderate value of $M = 6$.

\begin{figure}[t]
\centerline{\includegraphics[width=0.7\columnwidth]{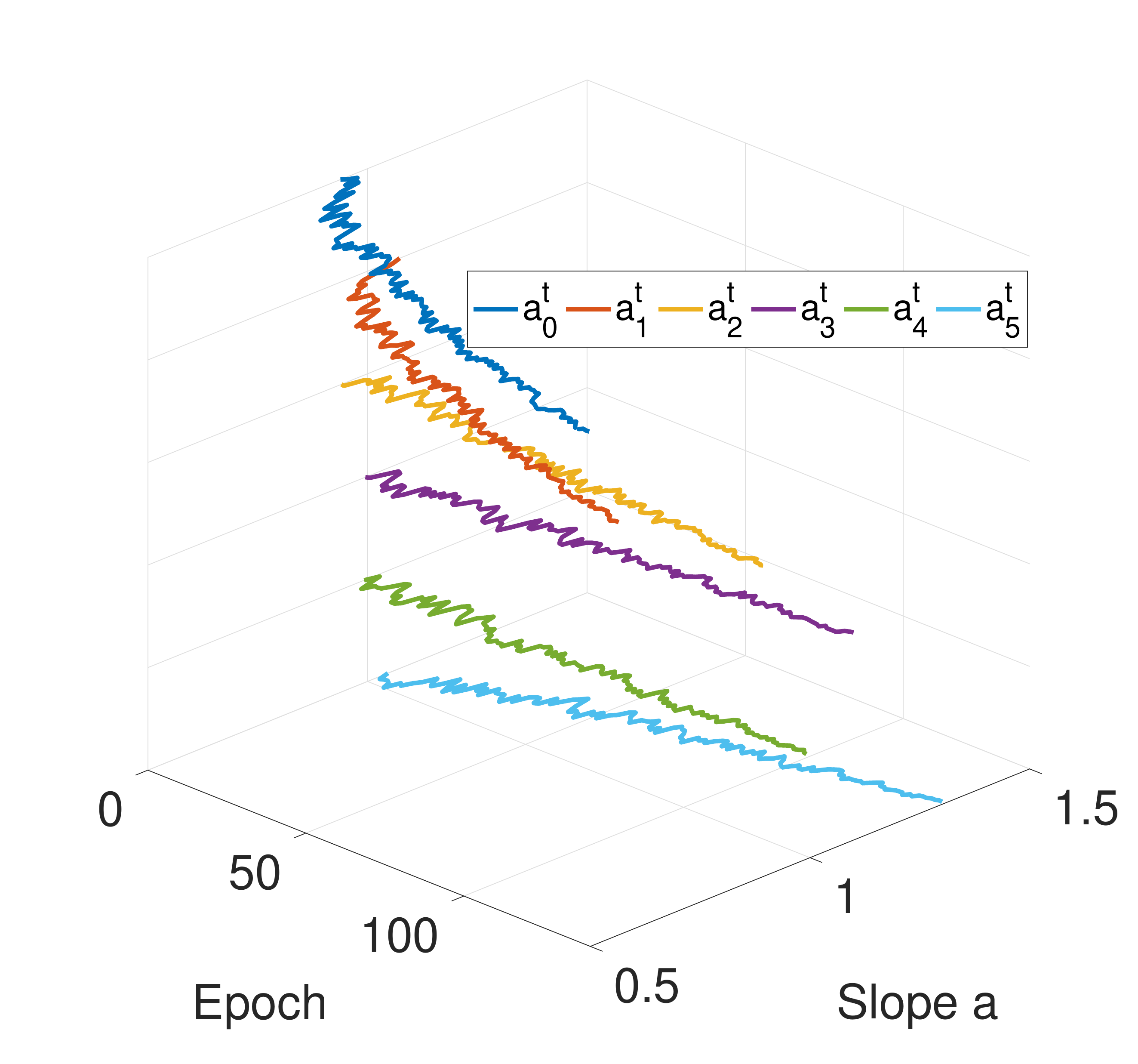}}
\caption{Convergency analysis of AM-LFS. 
}
\label{fig:converge}
\end{figure}

\textbf{Investigation on convergency:}
To evaluate the training process of the our AM-LFS, we need to conduct investigation on the training convergency especially at the outer level.
However, it's hard to simply track the loss convergency since our AM-LFS learns dynamic loss function distributions during the training process.
Tracking the average reward is also not a good idea because this signal is very noisy which would create the illusion of instability of the training progress. The main reason is that small updates to loss function distributions at outer level may lead to large changes to the network parameters at the inner level.
As a result, we choose to track a more intuitive metric, the distribution parameters to study the convergency of our method.
From the Figure~\ref{fig:converge}, we see that the distribution parameters tends to converge to specific values as the epochs increase, which indicates AM-LFS can be trained in a stable manner.

\begin{figure}[t]
\centerline{\includegraphics[width=0.7\columnwidth]{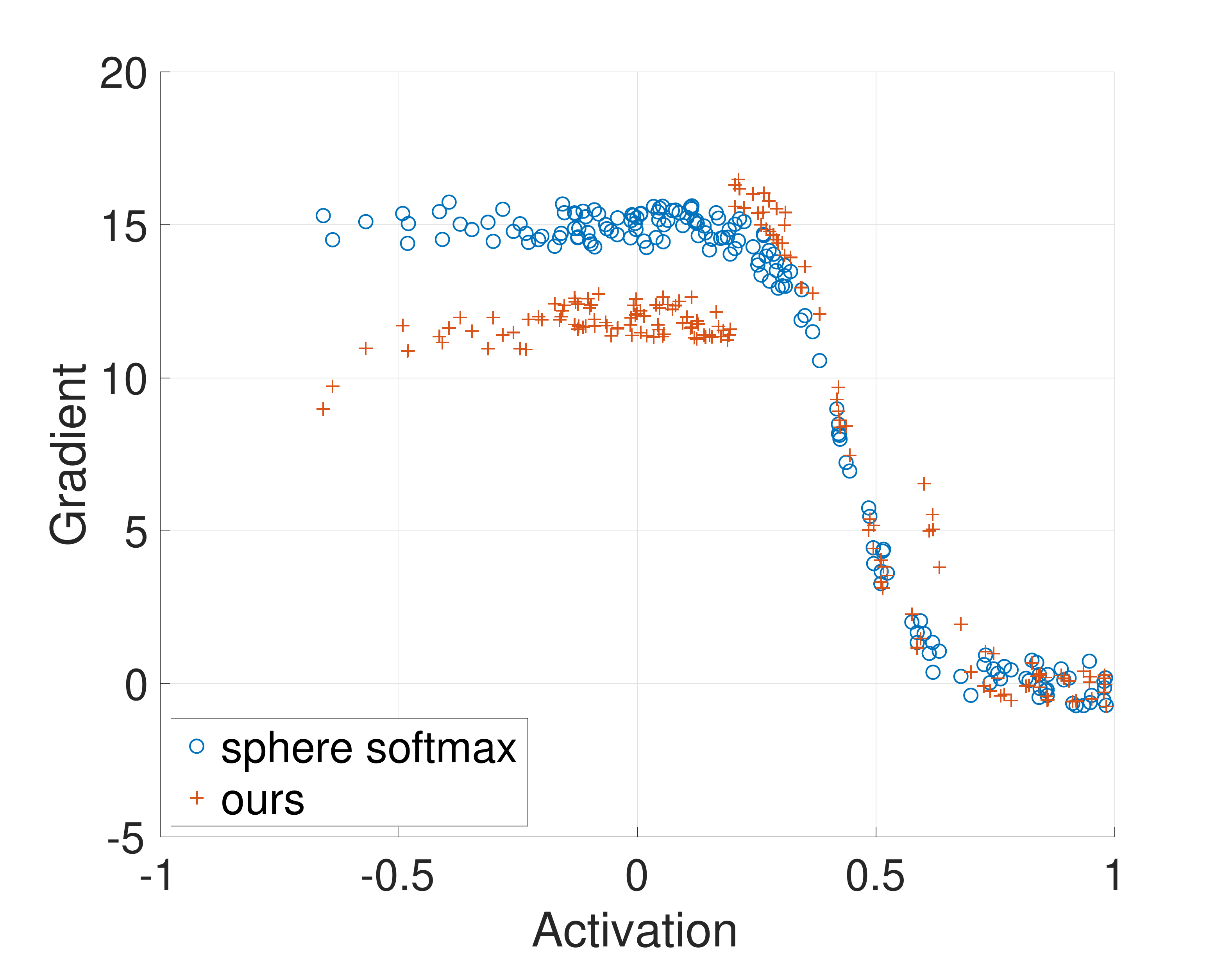}}
\caption{Visualization of gradient distribution of Sphere SoftmaxLoss and our AM-LFS.
}
\label{fig:grads}
\end{figure}

\textbf{Visualization of gradient distribution:}
We visualize the gradient distribution of intra-class distance term $f_{y_i}$ in Figure~\ref{fig:grads} to demonstrate AM-LFS that has more discrimination power than the baseline Sphere softmax loss function in SphereReID on Market-1501 dataset.
When the activation value (x-axis) increases, the intra distance will decrease, where data samples are at a relatively easy level.
On the contrary, when the activation value (x-axis) decreases, data samples are at a hard level.
As can be observed from Figure~\ref{fig:grads}, the gradients of AM-LFS with regard to hard examples are lower than those of baseline sphere softmax, which leads to a focus on the inter-class distance. We conclude that AM-LFS can dynamic tradeoff the significance of intra-class distance and inter-class distance hence boost the model's discrimination power.

\section{Conclusion}
In this paper, we have proposed AutoML for Loss Function Search (AM-LFS) which leverages REINFORCE to search loss functions during the training process. 
We carefully design an effective and task independent search space and bilevel optimization framework, which guarantees the generalization and transferability on different vision tasks.
While this paper only demonstrates the effectiveness of AM-LFS applying to classification, face recognition and person re-id datasets, it can also be easily applied to other off-the-shelf modern computer vision frameworks for various tasks, which is an interesting future work.

{\small
\bibliographystyle{ieee_fullname}
\bibliography{egbib}
}

\end{document}